\documentclass[sigconf]{acmart}

\usepackage{amsmath,amssymb}
\usepackage{graphicx}
\usepackage{float}
\usepackage{multirow}
\usepackage{booktabs}
\usepackage{grffile}
\usepackage{color}
\usepackage{xspace}
\usepackage{multirow}
\usepackage{hyperref}
\usepackage{subcaption}

\def\electricity{{\sf Electricity}\xspace}
\def\traffic{{\sf Traffic}\xspace}

\def\solar{{\sf Solar-Energy}\xspace}

\def\exchange{{\sf Exchange-Rate}\xspace}

\def\VARMLP{{\sf VAR-MLP}\xspace}
\def\LRidge{{\sf LRidge}\xspace}
\def\LSVR{{\sf LSVR}\xspace}
\def\GP{{\sf GP}\xspace}
\def\TRMF{{\sf TRMF}\xspace}
\def\AR{{\sf AR}\xspace}

\newcommand{\minimize}{\mathop{\mathrm{minimize}}}

\newcommand{\st}{\mathop{\mathrm{subject\,\,to}}}

\def\bc{{\boldsymbol c}}

\def\by{{\boldsymbol y}}

\def\bX{{\boldsymbol X}}
\def\bY{{\boldsymbol Y}}

\def\balpha{{\boldsymbol \alpha}}
\def\R{{\mathbb{R}}}

\setcopyright{rightsretained}

\acmDOI{10.475/123_4}

\acmISBN{123-4567-24-567/08/06}

\acmYear{2018}
\copyrightyear{2018}

\acmPrice{15.00}

\begin{document}
\title{Modeling Long- and Short-Term Temporal Patterns with Deep Neural Networks}

\author{Guokun Lai}
\affiliation{%
  \institution{Carnegie Mellon University}
}
\email{guokun@cs.cmu.edu}

\author{Wei-Cheng Chang}
\affiliation{%
  \institution{Carnegie Mellon University}
}
\email{wchang2@andrew.cmu.edu}

\author{Yiming Yang}
\affiliation{%
  \institution{Carnegie Mellon University}
}
\email{yiming@cs.cmu.edu}

\author{Hanxiao Liu}
\affiliation{%
  \institution{Carnegie Mellon University}
}
\email{hanxiaol@cs.cmu.edu}

\begin{abstract}
Multivariate time series forecasting is an important machine learning problem across many domains, including predictions of solar plant energy output, electricity consumption, and traffic jam situation. Temporal data arise in these real-world applications often involves a mixture of long-term and short-term patterns, for which traditional approaches such as Autoregressive models and Gaussian Process may fail. In this paper, we proposed a novel deep learning framework, namely Long- and Short-term Time-series network (LSTNet), to address this open challenge. LSTNet uses the Convolution Neural Network (CNN) and the Recurrent Neural Network (RNN) to extract short-term local dependency patterns among variables and to discover long-term patterns for time series trends. Furthermore, we leverage traditional autoregressive model to tackle the scale insensitive problem of the neural network model. In our evaluation on real-world data with complex mixtures of repetitive patterns, LSTNet achieved significant performance improvements over that of several state-of-the-art baseline methods. All the data and experiment codes are available online.

\end{abstract}


\maketitle

\section{Introduction}
\label{sec:intro}

Multivariate time series data are ubiquitous in our everyday life ranging from the prices in stock markets, the traffic flows on highways, the outputs of solar power plants, the temperatures across different cities, just to name a few. In such applications, users are often interested in the forecasting of the new trends or potential hazardous events based on historical observations on time series signals. For instance, a better route plan could be devised based on the predicted traffic jam patterns a few hours ahead, and a larger profit could be made with the forecasting of the near-future stock market. 

\begin{figure}[!t]
    	\centering
        \includegraphics[width=.45\textwidth]{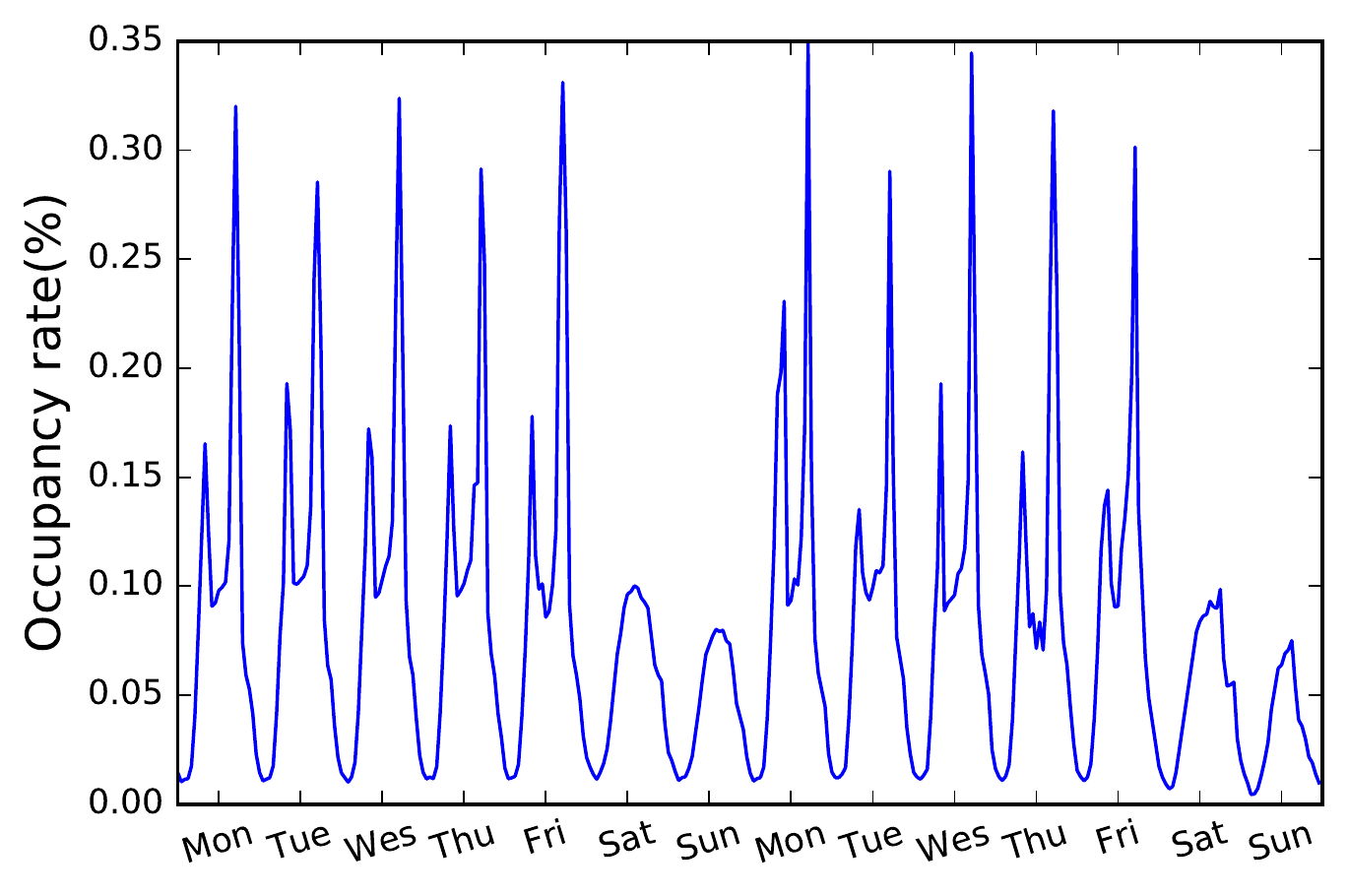}
        \caption{The hourly occupancy rate of a road in the bay area for 2 weeks}
        \label{fig:tra-ex}
    \end{figure}

Multivariate time series forecasting often faces a major research challenge, that is, how to capture and leverage the dynamics dependencies among multiple variables.  Specifically,  
real-world applications often entail a mixture of short-term and long-term repeating patterns, as shown in Figure \ref{fig:tra-ex} which plots the hourly occupancy rate of a freeway.
Apparently, there are two repeating patterns, daily and weekly. The former portraits the morning peaks vs. evening peaks, while the latter reflects the workday and weekend patterns. A successful time series forecasting model should be capture both kinds of recurring patterns for accurate predictions. As another example, consider the task of predicting the output of a solar energy farm based on the measured solar radiation by massive sensors over different locations.  The long-term patterns reflect the difference between days vs. nights, summer vs. winter, etc., and the short-term patterns reflect the effects of cloud movements, wind direction changes, etc.  Again, without taking both kinds of recurrent patterns into account, accurate time series forecasting is not possible.  However, traditional approaches such as the large body of work in autoregressive methods \cite{hamilton1994time, box2015time,zhang2003time,Yu_NIPS_16,li2014forecasting} fall short in this aspect, as most of them do not distinguish the two kinds of patterns nor model their interactions explicitly and dynamically. Addressing such limitations of existing methods in time series forecasting is the main focus of this paper, for which we propose a novel framework that takes advantages of recent developments in deep learning research.
    
    \begin{figure*}[!t]
    	\centering
        \includegraphics[width=\textwidth]{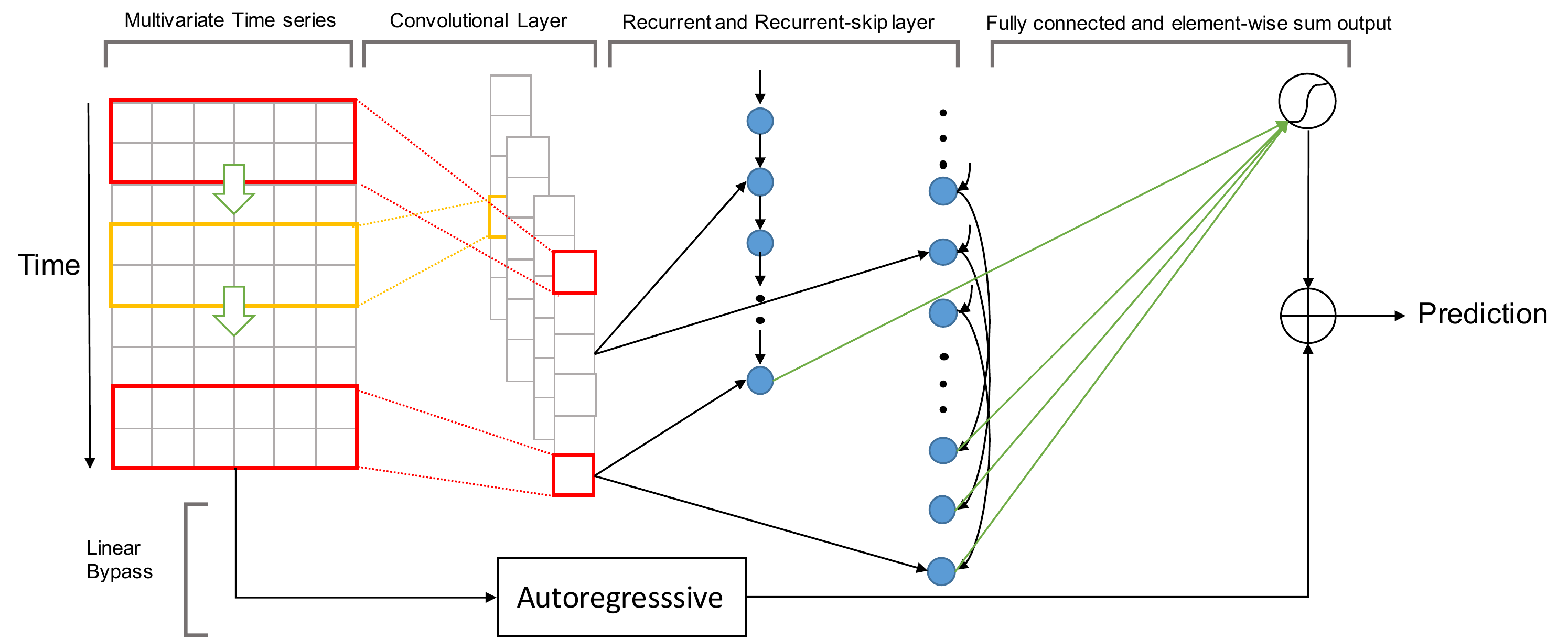}
        \caption{An overview of the Long- and Short-term Time-series network (LSTNet)}
        \label{fig:overview}
	\end{figure*}   
    
    Deep neural networks have been intensively studied in related domains, and made extraordinary impacts on the solutions of a broad range of problems.  The recurrent neural networks (RNN) models \cite{elman1990finding}, for example, have become most popular in recent natural language processing (NLP) research.
Two variants of RNN in particular, namely the Long Short Term Memory (LSTM) \cite{hochreiter1997long} and the Gated Recurrent Unit (GRU) \cite{chung2014empirical}, have significantly improved the state-of-the-art performance in machine translation, speech recognition and other NLP tasks as they can effectively capture the  meanings of words based on the long-term and short-term dependencies among them in input documents  \cite{bahdanau2014neural,hinton2012deep,krizhevsky2012imagenet}.
 In the field of computer vision, as another example, convolution neural network (CNN) models \cite{lecun1995convolutional, krizhevsky2012imagenet} have shown outstanding performance by successfully extracting local and shift-invariant features (called "shapelets" sometimes) at various granularity levels from input images.
    
Deep neural networks have also received an increasing amount of attention in time series analysis. A substantial portion of the previous work has been focusing on \textit{time series classification}, i.e., the task of automated assignment of class labels to time series input. For instance, RNN architectures have been studied for extracting informative patterns from health-care sequential data \cite{lipton2015learning,che2016recurrent} and classifying the data with respect diagnostic categories.  RNN has also been applied to mobile data, for classifying the input sequences with respect to actions or activities \cite{hammerla2016deep}. CNN models have also been used in action/activity recognition \cite{lea2016temporal,yang2015deep,hammerla2016deep}, for the extraction of shift-invariant local patterns from input sequences as the features of classification models.

Deep neural networks have also been studied for \textit{time series forecasting} \cite{dasgupta2016nonlinear, yu2017deep}, i.e., the task of using observed time series in the past to predict the unknown time series in a look-ahead horizon -- the larger the horizon, the harder the problem. Efforts in this direction range from the early work using naive RNN models \cite{connor1991recurrent} and the hybrid models \cite{zhang1998forecasting,zhang2003time,jain2007hybrid} combining the use of ARIMA \cite{box1970distribution} and Multilayer Perceptron (MLP), to the recent combination of vanilla RNN and Dynamic Boltzmann Machines in time series forecasting \cite{dasgupta2016nonlinear}.

In this paper, we propose a deep learning framework designed for the multivariate time series forecasting, namely Long- and Short-term Time-series Network (LSTNet), as illustrated in Figure \ref{fig:overview}. It leverages the strengths of both the convolutional layer to discover the local dependency patterns among multi-dimensional input variables and the recurrent layer to captures complex long-term dependencies. A novel recurrent structure, namely Recurrent-skip, is designed for capturing very long-term dependence patterns and making the optimization easier as it utilizes the periodic property of the input time series signals. Finally, the LSTNet incorporates a traditional autoregressive linear model in parallel to the non-linear neural network part, which makes the non-linear deep learning model more robust for the time series with violate scale changing. In the experiment on the real world seasonal time series datasets, our model consistently outperforms the traditional linear models and GRU recurrent neural network.

     The rest of this paper is organized as follows. Section \ref{sec:related} outlines the related background, including representative auto-regressive methods and Gaussian Process models. Section \ref{sec:model} describe our proposed LSTNet. Section \ref{sec:experiment} reports the evaluation results of our model in comparison with strong baselines on real-world datasets. Finally, we conclude our findings in Section \ref{sec:conclusion}.

\section{Related Background}
\label{sec:related}


One of the most prominent univariate time series models is the autoregressive integrated moving average (ARIMA) model. The popularity of the ARIMA model is due to its statistical properties as well as the well-known Box-Jenkins methodology \cite{box2015time} in the model selection procedure. ARIMA models are not only adaptive to various exponential smoothing techniques \cite{mckenzie1984general} but also flexible enough to subsume other types of time series models including autoregression (AR), moving average (MA) and Autoregressive Moving Average (ARMA). However, ARIMA models, including their variants for modeling long-term temporal dependencies \cite{box2015time}, are rarely used in high dimensional multivariate time series forecasting due to their high computational cost. 

On the other hand, vector autoregression (VAR) is arguably the most widely used models in multivariate time series \cite{hamilton1994time, box2015time, lutkepohl2005new} due to its simplicity. VAR models naturally extend AR models to the multivariate setting, which ignores the dependencies between output variables. Significant progress has been made in recent years in a variety of VAR models, including the elliptical VAR model \cite{qiu2015robust} for heavy-tail time series and structured VAR model \cite{melnyk2016estimating} for better interpretations of the dependencies between high dimensional variables, and more. Nevertheless, the model capacity of VAR grows linearly over the temporal window size and quadratically over the number of variables. This implies, when dealing with long-term temporal patterns, the inherited large model is prone to overfitting. To alleviate this issue, \cite{Yu_NIPS_16} proposed to reduce the original high dimensional signals into lower dimensional hidden representations, then applied VAR for forecasting with a variety choice of regularization.

Time series forecasting problems can also be treated as standard regression problems with time-varying parameters. It is therefore not surprising that various regression models with different loss functions and regularization terms are applied to time series forecasting tasks. For example, linear support vector regression (SVR) \cite{kim2003financial, cao2003support} learns a max margin hyperplane based on the regression loss with a hyper-parameter $\epsilon$ controlling the threshold of prediction errors. Ridge regression is yet another example which can be recovered from SVR models by setting $\epsilon$ to zeros. Lastly, \cite{li2014forecasting} applied LASSO models to encourage sparsity in the model parameters so that interesting patterns among different input signals could be manifest. These linear methods are practically more efficient for multivariate time series forecasting due to high-quality off-the-shelf solvers in the machine learning community. Nonetheless, like VARs, those linear models may fail to capture complex non-linear relationships of multivariate signals, resulting in an inferior performance at the cost of its efficiency.

Gaussian Processes (GP) is a non-parametric method for modeling distributions over a continuous domain of functions. This contrasts with models defined by a parameterized class of functions such as VARs and SVRs. GP can be applied to multivariate time series forecasting task as suggested in \cite{roberts2013gaussian}, and can be used as a prior over the function space in Bayesian inference. For example, \cite{frigola2013bayesian} presented a fully Bayesian approach with the GP prior for nonlinear state-space models, which is capable of capturing complex dynamical phenomena. However, the power of Gaussian Process comes with the price of high computation complexity. A straightforward implementation of Gaussian Process for multivariate time-series forecasting has cubic complexity over the number of observations, due to the matrix inversion of the kernel matrix.

\section{Framework}
\label{sec:model}

In this section, we first formulate the time series forecasting problem, and then discuss the details of the proposed LSTNet architecture (Figure \ref{fig:overview}) in the following part. Finally, we introduce the objective function and the optimization strategy.
\subsection{Problem Formulation}
\label{sec:format}

In this paper, we are interested in the task of multivariate time series forecasting. More formally, given a series of fully observed time series signals $\bY = \{\by_1, \by_2, \ldots, \by_T \}$ where $\by_t \in \R^n$, and $n$ is the variable dimension, we aim at predicting a series of future signals in a rolling forecasting fashion. That being said, to predict $\by_{T+h}$ where $h$ is the desirable horizon ahead of the current time stamp, we assume $\{ \by_1, \by_2, \ldots, \by_T \}$ are available. Likewise, to predict the  value of the next time stamp $\by_{T+h+1}$, we assume $\{ \by_1, \by_2, \ldots, \by_T, \by_{T+1}\}$ are available. We hence formulate the input matrix at time stamp $T$ as $X_T = \{ \by_1, \by_2, \ldots, \by_T\} \in \R^{n \times T}$. 

In the most of cases, the horizon of the forecasting task is chosen according to the demands of the environmental settings, e.g. for the traffic usage, the horizon of interest ranges from hours to a day; for the stock market data, even seconds/minutes-ahead forecast can be meaningful for generating returns.

Figure \ref{fig:overview} presents an overview of the proposed LSTnet architecture. The LSTNet is a deep learning framework specifically designed for multivariate time series forecasting tasks with a mixture of long- and short-term patterns. In following sections, we introduce the building blocks for the LSTNet in detail. 

\subsection{Convolutional Component}
The first layer of LSTNet is a convolutional network without pooling, which aims to extract short-term patterns in the time dimension as well as local dependencies between variables. The convolutional layer consists of multiple filters
of width $\omega$ and height $n$ (the height is set to be the same as the number of variables). The $k$-th filter sweeps through the input matrix $X$ and produces
\begin{equation}
h_k = RELU(W_k * X + b_k)
\end{equation}
where $*$ denotes the convolution operation and the output $h_k$ would be a vector, and the $RELU$ function is $RELU(x) = \max(0,x)$. We make each vector $h_k$ of length $T$ by zero-padding on the left of input matrix $X$. The output matrix of the convolutional layer is of size $d_c \times T$
where $d_c$ denotes the number of filters.

\subsection{Recurrent Component}
The output of the convolutional layer is simultaneously fed into the Recurrent component and Recurrent-skip component (to be described in subsection \ref{subsec:recurrent-skip}). The Recurrent component is a recurrent layer with the Gated Recurrent Unit (GRU) \cite{chung2014empirical} and uses the $RELU$ function as the hidden update activation function. The hidden state of recurrent units at time $t$ is computed as,

\begin{equation}
\begin{aligned}
r_t &= \sigma(x_tW_{xr} + h_{t-1}W_{hr} + b_r) \\
u_t &= \sigma(x_tW_{xu} + h_{t-1}W_{hu} + b_u) \\
c_t &= RELU(x_tW_{xc} + r_{t}\odot(h_{t-1}W_{hc}) + b_c) \\
h_t &= (1-u_t) \odot h_{t-1} + u_t \odot c_t \\
\end{aligned}
\label{eq:gru}
\end{equation}

where $\odot$ is the element-wise product, $\sigma$ is the sigmoid function and $x_t$ is the input of this layer at time $t$. The output of this layer is the hidden state at each time stamp. While researchers are accustomed to using $\tanh$ function as hidden update activation function, we empirically found $RELU$ leads to more reliable performance, through which the gradient is easier to back propagate. 

\subsection{Recurrent-skip Component}
\label{subsec:recurrent-skip}

The Recurrent layers with GRU \cite{chung2014empirical} and LSTM \cite{hochreiter1997long} unit are carefully designed to memorize the historical information and hence to be aware of relatively long-term dependencies. Due to gradient vanishing, however, GRU and LSTM usually fail to capture very long-term correlation in practice. We propose to alleviate this issue via a novel recurrent-skip component which leverages the periodic pattern in real-world sets. For instance, both the electricity consumption and traffic usage exhibit clear pattern on a daily basis. If we want to predict the electricity consumption at $t$ o'clock for today, a classical trick in the seasonal forecasting model is to leverage the records at $t$ o'clock in historical days, besides the most recent records.
This type of dependencies can hardly be captured by off-the-shelf recurrent units due to the extremely long length of one period (24 hours) and the subsequent optimization issues. Inspired by the effectiveness of this trick, we develop a recurrent structure with temporal skip-connections to extend the temporal span of the information flow and hence to ease the optimization process. Specifically, skip-links are added between the current hidden cell and the hidden cells in the same phase in adjacent periods. The updating process can be formulated as,

\begin{equation}
\begin{aligned}
r_t &= \sigma(x_tW_{xr} + h_{t-p}W_{hr} + b_r) \\
u_t &= \sigma(x_tW_{xu} + h_{t-p}W_{hu} + b_u) \\
c_t &= RELU(x_tW_{xc} + r_{t}\odot(h_{t-p}W_{hc}) + b_c) \\
h_t &= (1-u_t) \odot h_{t-p} + u_t \odot c_t \\
\end{aligned}
\label{eq:rnn-skip}
\end{equation}

where the input of this layer is the output of the convolutional layer, and $p$ is the number of hidden cells skipped through. The value of $p$ can be easily determined for datasets with clear periodic patterns (e.g. $p=24$ for the hourly electricity consumption and traffic usage datasets), and has to be tuned otherwise. In our experiments, we empirically found that a well-tuned $p$ can considerably boost the model performance even for the latter case. Furthermore, the LSTNet could be easily extended to contain variants of the skip length $p$.

We use a dense layer to combine the outputs of the Recurrent and Recurrent-skip components. The inputs to the dense layer include the hidden state of Recurrent component at time stamp $t$, denoted by $h^R_t$, and $p$ hidden states of Recurrent-skip component from time stamp $t-p+1$ to $t$ denoted by $h^S_{t-p+1},h^S_{t-p+2} \ldots, h^S_{t}$. The output of the dense layer is computed as,

\begin{equation}
h^D_t = W^R h^R_t + \sum_{i=0}^{p-1} W^S_{i}h^S_{t-i} + b
\label{eq:dense}
\end{equation}

where $h^D_t$ is the prediction result of the neural network (upper) part in the Fig.\ref{fig:overview} at time stamp $t$.

\subsection{Temporal Attention Layer}

However, the Recurrent-skip layer requires a predefined hyper-parameter $p$,
which is unfavorable in the nonseasonal time series prediction,
or whose period length is dynamic over time.
To alleviate such issue, 
we consider an alternative approach, attention mechanism \cite{bahdanau2014neural}, which learns the weighted combination of
hidden representations at each window position of the input matrix. Specifically, the attention weights $\balpha_t \in \mathbb{R}^q$ at current time stamp $t$ are calculated as
\begin{equation*}
	\balpha_{t} = AttnScore(H^R_t, h^R_{t-1})
\end{equation*}
where $H_t^R = [h^R_{t-q}, \ldots, h^R_{t-1}]$ is a matrix stacking the hidden representation
of RNN column-wisely and $AttnScore$ is some similarity functions such as dot product, cosine,
or parameterized by a simple multi-layer perceptron.

The final output of temporal attention layer is the concatenation of the weighted context vector $\bc_t = H_t \balpha_t$ and last window hidden representation $h^R_{t-1}$, along with
a linear projection operation
\begin{equation*}
h^D_t = W[\bc_t; h^R_{t-1}] + b.
\end{equation*}

\subsection{Autoregressive Component}
\label{sec:AR}
Due to the non-linear nature of the Convolutional and Recurrent components, one major drawback of the neural network model is that the scale of outputs is not sensitive to the scale of inputs. Unfortunately, in specific real datasets, the scale of input signals constantly changes in a non-periodic manner, which significantly lowers the forecasting accuracy of the neural network model. A concrete example of this failure is given in Section \ref{sec:ablation}. To address this deficiency, similar in spirit to the highway network \cite{srivastava2015highway}, we decompose the final prediction of LSTNet into a linear part, which primarily focuses on the local scaling issue, plus a non-linear part containing recurring patterns. In the LSTNet architecture, we adopt the classical Autoregressive (AR) model as the linear component. Denote the forecasting result of the AR component as $h^L_{t} \in \R^n$, and the coefficients of the AR model as $W^{ar} \in \R^{q^{ar}}$ and $b^{ar} \in \R$, where $q^{ar}$ is the size of input window over the input matrix. Note that in our model, all dimensions share the same set of linear parameters. The AR model is formulated as follows,
\begin{equation}
h^L_{t,i} = \sum_{k=0}^{q^{ar}-1}W^{ar}_k\by_{t-k,i} + b^{ar}
\end{equation}

The final prediction of LSTNet is then obtained by by integrating the outputs of the neural network part and the AR component:
\begin{equation}
\hat{\bY}_t = h^D_t + h^L_t
\end{equation}
where $\hat{\bY}_t$ denotes the model's final prediction at time stamp $t$.

\subsection{Objective function}
The squared error is the default loss function for many forecasting tasks,
the corresponding optimization objective is formulated as,
\begin{equation}
\minimize_{\Theta}~~~\sum_{t \in \Omega_{Train}} ||\bY_t - \hat{\bY}_{t-h}||_F^2
\label{eq:L2loss}
\end{equation}
where $\Theta$ denotes the parameter set of our model, $\Omega_{Train}$ is the set of time stamps used for training, $||\cdot||_F$ is the Frobenius norm, and $h$ is the horizon as mentioned in Section \ref{sec:format}.
The traditional linear regression model with the square loss function is named as Linear Ridge, which is equivalent to the vector autoregressive model with ridge regularization. However, experiments show that the Linear Support Vector Regression (Linear SVR) \cite{vapnik1997support} dominates the Linear Ridge model in certain datasets. The only difference between Linear SVR and Linear Ridge is the objective function. The objective function for Linear SVR is,

\begin{equation}
\begin{aligned}
\minimize_{\Theta}~~~&\frac{1}{2}||\Theta||_F^2 + C\sum_{t \in \Omega_{Train}} \sum_{i=0}^{n-1} \xi_{t,i} \\
\st~~~& |\hat{\bY}_{t-h,i} - \bY_{t,i}| \le \xi_{t,i} + \epsilon, t \in \Omega_{Train} \\
& \xi_{t,i} \ge 0
\end{aligned}
\label{eq:svr}
\end{equation}
where $C$ and $\epsilon$ are hyper-parameters. Motivated by the remarkable performance of the Linear SVR model, we incorporate its objective function in the LSTNet model as an alternative of the squared loss. For simplicity, we assume $\epsilon=0$\footnote{One could keep $\epsilon$ to make the objective function more faithful to the Linear SVR model without modifying the optimization strategy. We leave this for future study.}, and the objective function above reduces to absolute loss (L1-loss) function as follows:
\begin{equation}
\minimize_{\Theta}~~~\sum_{t \in \Omega_{Train}} \sum_{i=0}^{n-1}|\bY_{t,i} - \hat{\bY}_{t-h,i}|
\label{eq:L1loss}
\end{equation}

The advantage of the absolute loss function is that it is more robust to the anomaly in the real time series data. In the experiment section, we use the validation set to decide to use which objective function, square loss Eq.\ref{eq:L2loss} or absolute one Eq.\ref{eq:L1loss}.

\subsection{Optimization Strategy}
\label{sec:train}
In this paper, our optimization strategy is the same as that in the traditional time series forecasting model. Supposing the input time series is $\bY_t = \{\by_1, \by_2, \ldots, \by_t \}$, we define a tunable window size $q$, and reformulate the input at time stamp $t$ as $\bX_t = \{\by_{t-q+1}, \by_{t-q+2}, \ldots, \by_t \}$. The problem then becomes a regression task with a set of feature-value pairs $\{\bX_t, \bY_{t+h}\}$, and can be solved by Stochastic Gradient Decent (SGD) or its variants such as Adam \cite{kingma2014adam}.

\section{Evaluation}
\label{sec:experiment}


We conducted extensive experiments with 9 methods (including our new methods) on 4 benchmark datasets for time series forecasting tasks. All the data and experiment codes are available online \footnote{the link is anonymous due to the double blind policy}. 

\subsection{Methods for Comparison}
\label{sec:baseline}
The methods in our comparative evaluation are the follows.
\begin{itemize}
    \item \AR stands for the autoregressive model, which is equivalent to the one dimensional VAR model. 
    \item \LRidge is the vector autoregression (VAR) model with L2-regularization, which has been most popular for multivariate time series forecasting.
    \item \LSVR is the vector autoregression (VAR) model with Support Vector Regression objective function \cite{vapnik1997support} . 
    \item \TRMF is the autoregressive model using temporal regularized matrix factorization by \cite{Yu_NIPS_16}.
    \item \GP is the Gaussian Process for time series modeling. \cite{frigola2015bayesian,roberts2013gaussian}
    \item \VARMLP is the model proposed in \cite{zhang2003time} that combines Multilayer Perception (MLP) and autoregressive model. 
    \item RNN-GRU is the Recurrent Neural Network model using GRU cell. 
    \item LSTNet-skip is our proposed LSTNet model with skip-RNN layer. 
    \item LSTNet-Attn is our proposed LSTNet model with temporal attention layer.
\end{itemize}
For the single output methods above such as \AR, \LRidge, \LSVR and \GP, we just trained $n$ models independently, i.e., one model for each of the $n$ output variables.

\subsection{Metrics}
\label{sec:metrics} 
We used three conventional evaluation metrics defined as:
\begin{itemize}
	\item Root Relative Squared Error (RSE):
        \begin{equation}
        \begin{aligned}
        RSE = \frac{\sqrt{\sum_{(i,t) \in \Omega_{Test}}(Y_{it} - \hat{Y}_{it})^2}}
        			{\sqrt{\sum_{(i,t) \in \Omega_{Test}} (Y_{it} - mean(\bY))^2}}
        \end{aligned}
        \label{eq:RSE}
        \end{equation}
    \item Empirical Correlation Coefficient (CORR)
        \begin{equation}
        CORR =\frac{1}{n} \sum_{i=1}^n \frac{\sum_t \big(Y_{it} - mean({\bY}_i)\big) \big(\hat{Y}_{it} - mean(\hat{\bY}_i)\big)}
        	{\sqrt{\sum_t \big(Y_{it} - mean({\bY}_i)\big)^2 \big(\hat{Y}_{it} - mean(\hat{\bY}_i)\big)^2}}
        \label{eq:Correlation}
        \end{equation}    
\end{itemize}
where $\bY,\hat{\bY} \in \mathbb{R}^{n\times T}$ are ground true signals and system prediction signals, respectively. 
The RSE are the scaled version of the widely used Root Mean Square Error(RMSE), which is design to make more readable evaluation, regardless the data scale. 
For RSE lower value is better, while for CORR higher value is better.

\subsection{Data}
\label{sec:data}
We used four benchmark datasets which are publicly available.
Table \ref{tb:data-stats} summarizes the corpus statistics.
\begin{itemize}
    \item \traffic \footnote{\url{http://pems.dot.ca.gov}}: A collection of 48 months (2015-2016) hourly data from the California Department of Transportation. The data describes the road occupancy rates (between 0 and 1) measured by different sensors on San Francisco Bay area freeways.
    \item \solar \footnote{\url{http://www.nrel.gov/grid/solar-power-data.html}} : the solar power production records in the year of 2006, which is sampled every 10 minutes from 137 PV plants in Alabama State. 
	\item \electricity \footnote{\url{https://archive.ics.uci.edu/ml/datasets/ElectricityLoadDiagrams20112014}}: The electricity consumption in kWh was recorded every 15 minutes from 2012 to 2014, for n = 321 clients. We converted the data to reflect hourly consumption; 
    \item \exchange: the collection of the daily exchange rates of eight foreign countries including Australia, British, Canada, Switzerland, China, Japan, New Zealand and Singapore ranging from 1990 to 2016.
\end{itemize}

\begin{table}
	\begin{center}
        \resizebox{0.75\columnwidth}{!}{%
		\begin{tabular}{lrrrrrrrr} 
		\toprule
		Datasets 		& $T$		& $D$	& $L$		 \\
		\midrule
		\traffic 		& 17,544 	& 862 	& 1 hour	\\
        \solar			& 52,560	& 137	& 10 minutes\\
		\electricity 	& 26,304 	& 321 	& 1 hour	\\
        \exchange		& 7,588		& 8		& 1 day		\\
        \bottomrule
		\end{tabular}
        }
        \caption{Dataset Statistics, where $T$ is length of time series, $D$ is number of variables, $L$ is the sample rate.}
		\label{tb:data-stats}
	\end{center}
\end{table}

All datasets have been split into training set (60\%), validation set (20\%) and test set (20\%) in chronological order. To facilitate future research in multivariate time series forecasting, we publicize all raw datasets and the one after preprocessing in the website.

\begin{figure}[!ht]
\begin{subfigure}{.23\textwidth}
  \centering
  \includegraphics[width=\linewidth]{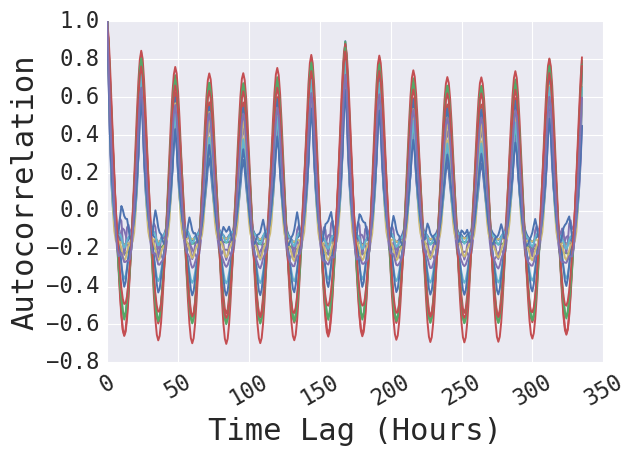}
  \caption{\traffic dataset}
\end{subfigure}
\begin{subfigure}{.23\textwidth}
  \centering
  \includegraphics[width=\linewidth]{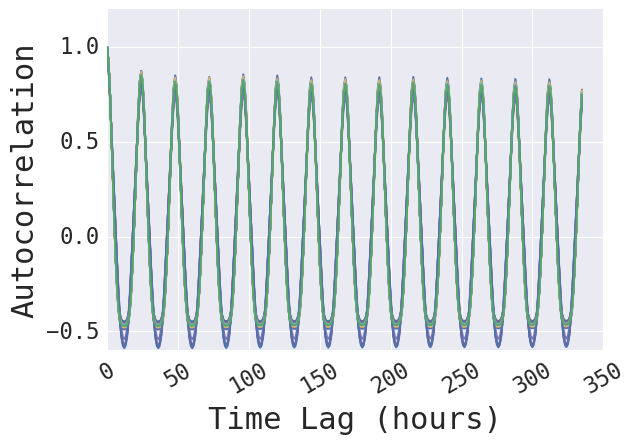}
  \caption{\solar dataset}
\end{subfigure}
\begin{subfigure}{.23\textwidth}
  \centering
  \includegraphics[width=\linewidth]{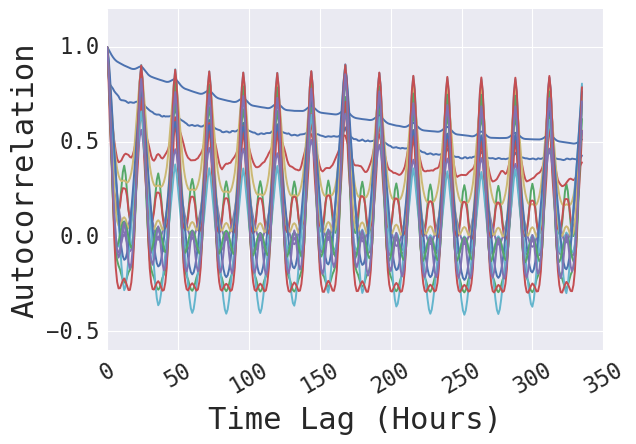}
  \caption{\electricity dataset}
\end{subfigure}
\begin{subfigure}{.23\textwidth}
  \centering
  \includegraphics[width=\linewidth]{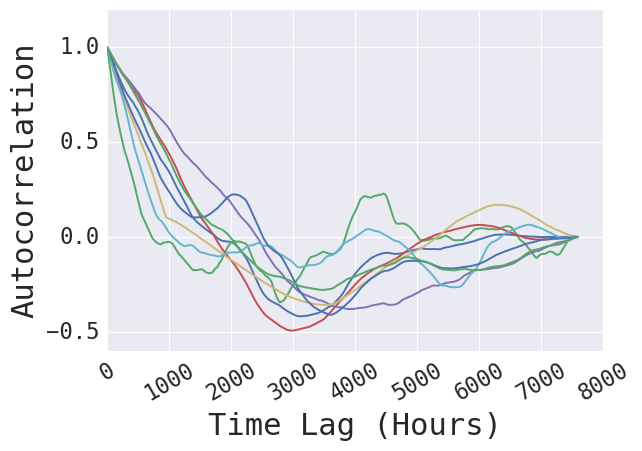}
  \caption{\exchange dataset}
\end{subfigure}
\caption{Autocorrelation graphs of sampled variables form four datasets.}
\label{fig:autocorrelation}
\end{figure}

In order to examine the existence of long-term and/or short-term repetitive patterns in time series data, we plot autocorrelation graph for some randomly selected variables from the four datasets in Figure \ref{fig:autocorrelation}. Autocorrelation, also known as serial correlation, is the correlation of a signal with a delayed copy of itself as a function of delay defined below
\begin{equation*}
	R(\tau) =  \frac{\mathbb{E}[(X_t - \mu)(X_{t+\tau} - \mu)]}{\sigma^2}
\end{equation*}
where $X_t$ is the time series signals, $\mu$ is mean and $\sigma^2$ is variance. In practice, we consider the empirical unbiased estimator to calculate the autocorrelation.  

We can see in the graphs (a), (b) and (c) of Figure \ref{fig:autocorrelation}, there are repetitive patterns with high autocorrelation in the \traffic, \solar and \electricity datasets, but not in the \exchange dataset. Furthermore, we can observe a short-term daily pattern (in every 24 hours) and long-term weekly pattern (in every 7 days) in the graph of the \traffic and \electricity dataset, which perfect reflect the expected regularity in highway traffic situations and electricity consumptions.  On the other hand, in graph (d) of the \exchange dataset, we hardly see any repetitive long-term patterns, expect some short-term local continuity. These observations are important for our later analysis on the empirical results of different methods.  That is, for the methods which can properly model and successfully leverage both short-term and long-term repetitive patterns in data, they should outperform well when the data contain such repetitive patterns (like in \electricity, \traffic and \solar).  On the other hand, if the dataset does not contain such patterns (like in \exchange), the advantageous power of those methods may not lead a better performance than that of other less powerful methods.  We will revisit this point in Section \ref{sec:mixture} with empirical justifications.

\begin{table*}[!ht]
\centering
\resizebox{\textwidth}{!}{
\begin{tabular}{ll|cccc|cccc|cccc|cccc}
\toprule
Dataset &  & \multicolumn{4}{c|}{\solar} & \multicolumn{4}{c|}{\traffic} & \multicolumn{4}{c|}{\electricity} & \multicolumn{4}{c}{\exchange}  \\
\midrule
										 		&         & \multicolumn{4}{c|}{Horizon} & \multicolumn{4}{c|}{Horizon} & \multicolumn{4}{c|}{Horizon} & \multicolumn{4}{c}{Horizon} \\
\midrule
Methods 	& Metrics & 3 & 6 & 12 & 24 & 3 & 6 & 12 & 24 & 3 & 6 & 12 & 24 & 3 & 6 & 12 & 24 \\
\midrule
\multirow{1}{*}{\AR} & RSE 
                          & 0.2435 & 0.3790 & 0.5911 & 0.8699 
                          & 0.5991 & 0.6218 & 0.6252 & 0.6293
                          & 0.0995 & 0.1035 & 0.1050 & 0.1054 
                          & 0.0228 & 0.0279 & \textbf{0.0353} & \textbf{0.0445} \\
\textbf{(3)}               
           & CORR 
                  & 0.9710 & 0.9263 & 0.8107 & 0.5314 
                  & 0.7752 & 0.7568 & 0.7544 & 0.7519
                  & 0.8845 & 0.8632 & 0.8591 & 0.8595 
                  & 0.9734 & 0.9656 & 0.9526 & \textbf{0.9357} \\   
\midrule
\multirow{1}{*}{LRidge} & RSE 
						              & 0.2019 & 0.2954 & 0.4832 & 0.7287 
                          & 0.5833 & 0.5920 & 0.6148 & 0.6025
                          & 0.1467 & 0.1419 & 0.2129 & 0.1280 
                          & \textbf{0.0184} & 0.0274 & 0.0419 & 0.0675\\
\textbf{(3)}              & CORR 
                   		  & 0.9807 & 0.9568 & 0.8765 & 0.6803 
                          & 0.8038 & 0.8051 & 0.7879 & 0.7862
                          & 0.8890 & 0.8594 & 0.8003 & 0.8806 
                          & \textbf{0.9788} & \textbf{0.9722} & 0.9543 & 0.9305 \\   
\midrule
\multirow{1}{*}{LSVR} & RSE 
							& 0.2021 & 0.2999 & 0.4846 & 0.7300 
                            & 0.5740 & 0.6580 & 0.7714 & 0.5909
                            & 0.1523 & 0.1372 & 0.1333 & 0.1180 
                            & 0.0189 & 0.0284 & 0.0425 & 0.0662 \\
\textbf{(1)}                & CORR 
                     		& 0.9807 & 0.9562 & 0.8764 & 0.6789 
                            & 0.7993 & 0.7267 & 0.6711 & 0.7850 
                            & 0.8888 & 0.8861 & 0.8961 & 0.8891 
                            & 0.9782 & 0.9697 & \textbf{0.9546} & 0.9370 \\   
\midrule
\multirow{1}{*}{TRMF} & RSE 
							& 0.2473 & 0.3470 & 0.5597 & 0.9005 
                            & 0.6708 & 0.6261 & 0.5956 & 0.6442
                            & 0.1802 & 0.2039 & 0.2186 & 0.3656 
                            & 0.0351 & 0.0875 & 0.0494 & 0.0563 \\
\textbf{(0)}         & CORR 
                     		& 0.9703 & 0.9418 & 0.8475 & 0.5598
                            & 0.6964 & 0.7430 & 0.7748 & 0.7278 
                            & 0.8538 & 0.8424 & 0.8304 & 0.7471 
                            & 0.9142 & 0.8123 & 0.8993 & 0.8678 \\
\midrule
\multirow{1}{*}{GP} & RSE 
						  & 0.2259 & 0.3286 & 0.5200 & 0.7973
                          & 0.6082 & 0.6772 & 0.6406 & 0.5995
                          & 0.1500 & 0.1907 & 0.1621 & 0.1273 
                          & 0.0239 & \textbf{0.0272} & 0.0394 & 0.0580 \\
\textbf{(1)}              & CORR 
                   		  & 0.9751 & 0.9448 & 0.8518 & 0.5971
                          & 0.7831 & 0.7406 & 0.7671 & 0.7909
                          & 0.8670 & 0.8334 & 0.8394 & 0.8818
                          & 0.8713 & 0.8193 & 0.8484 & 0.8278\\
\midrule
{\multirow{1}{*}{VARMLP}} & RSE 
						 	   & 0.1922 & 0.2679 & 0.4244 & 0.6841 
                               & 0.5582 & 0.6579 & 0.6023 & 0.6146 
                               & 0.1393 & 0.1620 & 0.1557 & 0.1274
                               & 0.0265 & 0.0304 & 0.0407 & 0.0578\\
\textbf{(0)}                   & CORR 
                        	   & 0.9829 & 0.9655 & 0.9058 & 0.7149 
                               & 0.8245 & 0.7695 & 0.7929 & 0.7891
                               & 0.8708 & 0.8389 & 0.8192 & 0.8679
                               & 0.8609 & 0.8725 & 0.8280 & 0.7675 \\
\midrule 
\multirow{1}{*}{RNN-GRU}	& RSE 
							  & 0.1932 & 0.2628 & 0.4163 & 0.4852
							  & 0.5358 & 0.5522 & 0.5562 & 0.5633
                			  & 0.1102 & 0.1144 & 0.1183 & 0.1295
							  & 0.0192 & 0.0264 & 0.0408 & 0.0626 \\
\textbf{(0)}               & CORR 
                       		  & 0.9823 & 0.9675 & 0.9150 & 0.8823 
                              & 0.8511 & 0.8405 & 0.8345 & 0.8300
                              & 0.8597 & 0.8623 & 0.8472 & 0.8651
                              & 0.9786 & 0.9712 & 0.9531 & 0.9223 \\ 
                              
\midrule                                        
\midrule
\multirow{1}{*}{LST-Skip}	& RSE 
							  & 0.1843 & 0.2559 & \textbf{0.3254} & 0.4643
							  & \textbf{0.4777} & \textbf{0.4893} & \textbf{0.4950} & \textbf{0.4973}
                			  & \textbf{0.0864} & \textbf{0.0931} & 0.1007 & \textbf{0.1007}
							  & 0.0226 & 0.0280 & 0.0356 & 0.0449\\
\textbf{(17)}               & CORR 
                       		  & 0.9843 & 0.9690 & \textbf{0.9467} & 0.8870 
                              & \textbf{0.8721} & \textbf{0.8690} & \textbf{0.8614} & \textbf{0.8588}
                              & \textbf{0.9283} & \textbf{0.9135} & \textbf{0.9077} & \textbf{0.9119}
                              & 0.9735 & 0.9658 & 0.9511 & 0.9354\\ 
                              
\midrule                                                
\multirow{1}{*}{LST-Attn}	& RSE 
							  & \textbf{0.1816} & \textbf{0.2538} & 0.3466 & \textbf{0.4403}
							  & 0.4897 & 0.4973 & 0.5173 & 0.5300
                			  & 0.0868 & 0.0953 & \textbf{0.0984} & 0.1059
							  & 0.0276 & 0.0321 & 0.0448 & 0.0590 \\
\textbf{(7)}               & CORR 
                       		  & \textbf{0.9848} & \textbf{0.9696} & 0.9397 & \textbf{0.8995} 
                              & 0.8704 & 0.8669 & 0.8540 & 0.8429
                              & 0.9243 & 0.9095 & 0.9030 & 0.9025
                              & 0.9717 & 0.9656 & 0.9499 & 0.9339 \\ 
                                                
\bottomrule
\end{tabular}
}
\caption{Results summary (in RSE and CORR) of all methods on four datasets: 1) each row has the results of a specific method in a particular metric; 2) each column compares the results of all methods on a particular dataset with a specific horizon value; 3) bold face indicates the best result of each column in a particular metric; and 4) the total number of bold-faced results of each method is listed under the method name within parentheses. }
\label{tb:result}
\end{table*}

\subsection{Experimental Details}
We conduct grid search over all tunable hyper-parameters on the held-out validation set for each method and dataset. Specifically, all methods share the same grid search range of the window size $q$ ranging from $\{2^0,2^1,\ldots,2^9\}$ if applied. For \LRidge and \LSVR, the regularization coefficient $\lambda$ is chosen from $\{2^{-10}, 2^{-8}, \ldots, 2^{8}, 2^{10}\}$. For \GP, the RBF kernel bandwidth $\sigma$ and the noise level $\alpha$ are chosen from $\{2^{-10}, 2^{-8}, \ldots, 2^{8}, 2^{10}\}$. For \TRMF, the hidden dimension is chosen from $\{2^2, \ldots, 2^6\}$ and the regularization coefficient $\lambda$ is chosen from $\{0.1, 1, 10\}$.  For LST-Skip and LST-Attn, we adopted the training strategy described in Section \ref{sec:train}. The hidden dimension of the Recurrent and Convolutional layer is chosen from $\{50,100,200\}$, and $\{20,50,100\}$ for Recurrent-skip layer. The skip-length $p$ of Recurrent-skip layer is set as 24 for the \traffic and \electricity dataset, and tuned range from $2^1$ to $2^6$ for the \solar and \exchange datasets. The regularization coefficient of the AR component is chosen from $\{0.1,1,10\}$ to achieve the best performance. We perform dropout after each layer, except input and output ones, and the rate usually is set to 0.1 or 0.2. The Adam\cite{kingma2014adam} algorithm is utilized to optimize the parameters of our model.

\subsection{Main Results}
\label{sec:result}

Table \ref{tb:result} summarizes the evaluation results of all the methods (8) on all the test sets (4) in all the metrics (3). We set $horizon = \{3,6,12,24\}$, respectively, which means the horizons was set from 3 to 24 hours for the forecasting over the \electricity and \traffic data, from 30 to 240 minutes over the \solar data, and from 3 to 24 days over the \exchange data. The larger the horizons, the harder the prediction tasks. The best result for each (data, metric) pair is highlighted in bold face in this table.  The total count of the bold-faced results is 17 for LSTNet-Skip (one version of the proposed LSTNet), 7 for LSTNet-Attn (the other version of our LSTNet), and between 0 to 3 for the rest of the methods.

Clearly, the two proposed models, LSTNet-skip and LSTNet-Attn, consistently enhance over state-of-the-art on the datasets with periodic pattern, especially in the settings of large horizons. 
Besides, LSTNet outperforms the strong neural baseline RNN-GRU by \textbf{9.2\%, 11.7\%, 22.2\%} in RSE metric
on \solar, \traffic and \electricity dataset respectively when the horizon is 24,
demonstrating the effectiveness of the framework design for complex repetitive patterns. 
What's more, when the periodic pattern $q$ is not clear from applications, users may consider LSTNet-attn as alternative over LSTNet-skip, given the former still yield considerable improvement
over the baselines. But the proposed LSTNet is slightly worse than AR and LRidge on the \exchange dataset. Why?  Recall that in Section \ref{sec:data} and Figure  \ref{fig:autocorrelation} we used the autocorrelation curves of these datasets to show the existence of  repetitive patterns in the \solar, \traffic and \electricity datasets but not in \exchange.  The current results provide empirical evidence for the success of LSTNet models in modeling long-term and short-term dependency patterns when they do occur in data.  Otherwise, LSTNet performed comparably  with the better ones (AR and LRidge) among the representative baselines.  


Compared the results of univariate AR with that of the multivariate baseline methods (LRidge, LSVR and RNN), we see that in some datasets, i.e. \solar and \traffic, the multivariate approaches is stronger, but weaker otherwise, which means that the richer input information would causes overfitting in the traditional multivariate approaches. In contrast, the LSTNet has robust performance in different situations, partly due to its  autoregressive component, which we will discuss further in Section \ref{sec:ablation}.

\subsection{Ablation Study}
\label{sec:ablation}
To demonstrate the efficiency of our framework design, a careful ablation study is conducted. Specifically, we remove each component one at a time in our LSTNet framework. First, we name the LSTNet without different components as follows.

\begin{itemize}
\item \textbf{LSTw/oskip}: The LSTNet models without the Recurrent-skip component and attention component.
\item \textbf{LSTw/oCNN}: The LSTNet-skip models without the Convolutional component.
\item \textbf{LSTw/oAR}: The LSTNet-skip models  without the AR component.
\end{itemize}

For different baselines, we tune the hidden dimension of models such that they have similar numbers of model parameters to the completed LSTNet model, removing the performance gain induced by model complexity.

\begin{figure*}[!ht]

\begin{subfigure}{\textwidth}
  \centering
  \includegraphics[width=.45\linewidth]{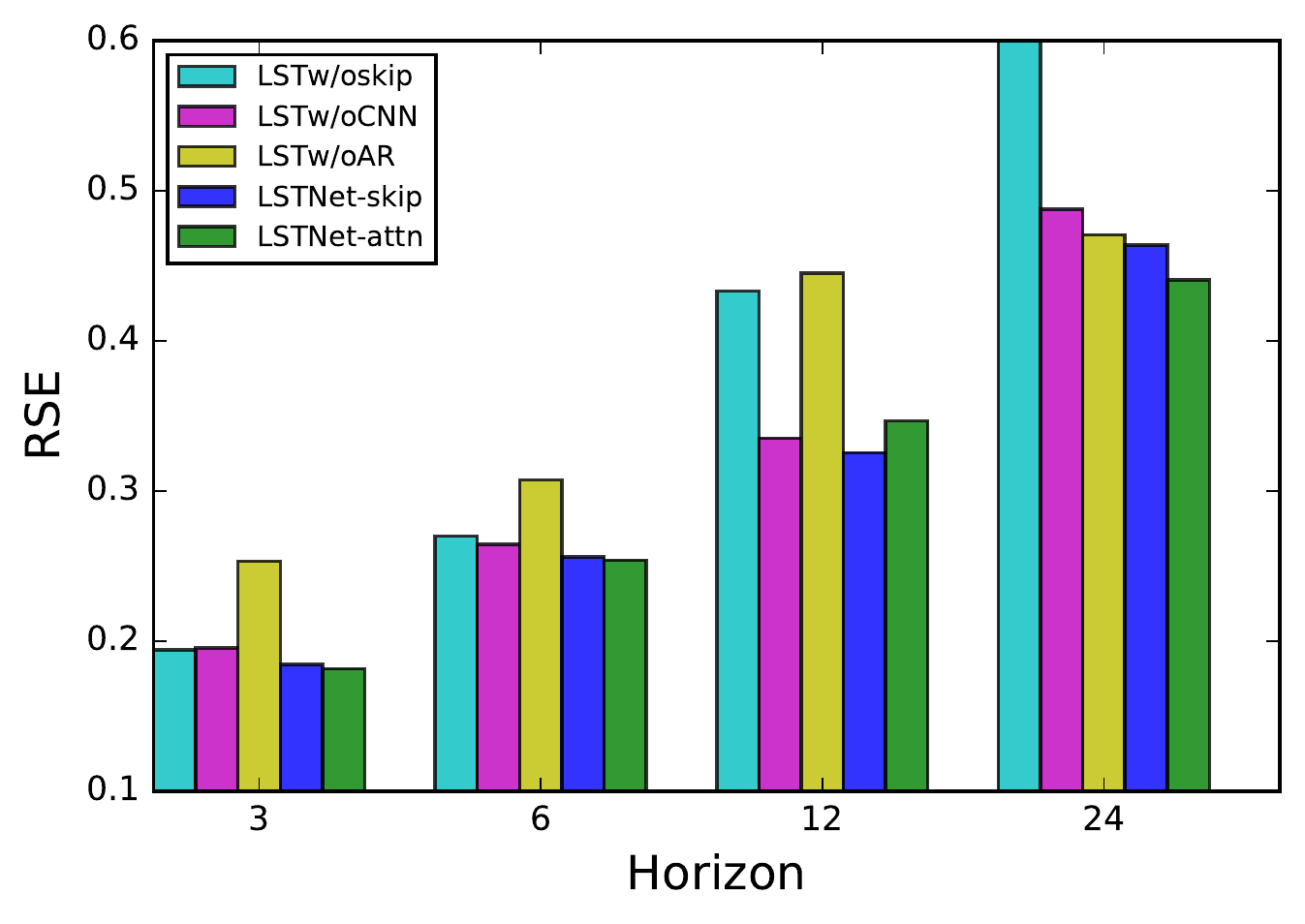}
  \includegraphics[width=.45\linewidth]{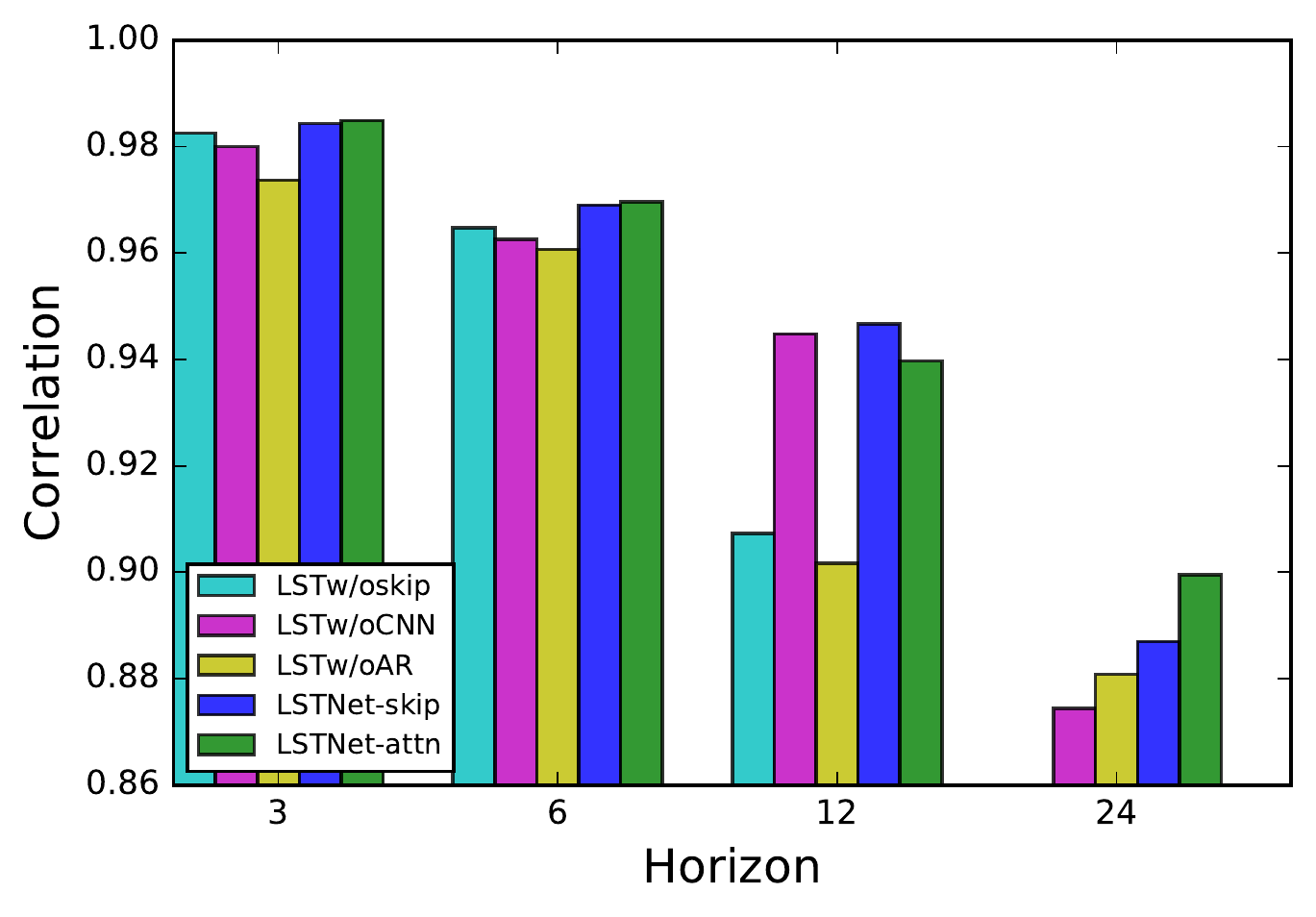}
  \caption{\solar dataset}
\end{subfigure}
\begin{subfigure}{\textwidth}
  \centering
  \includegraphics[width=.45\linewidth]{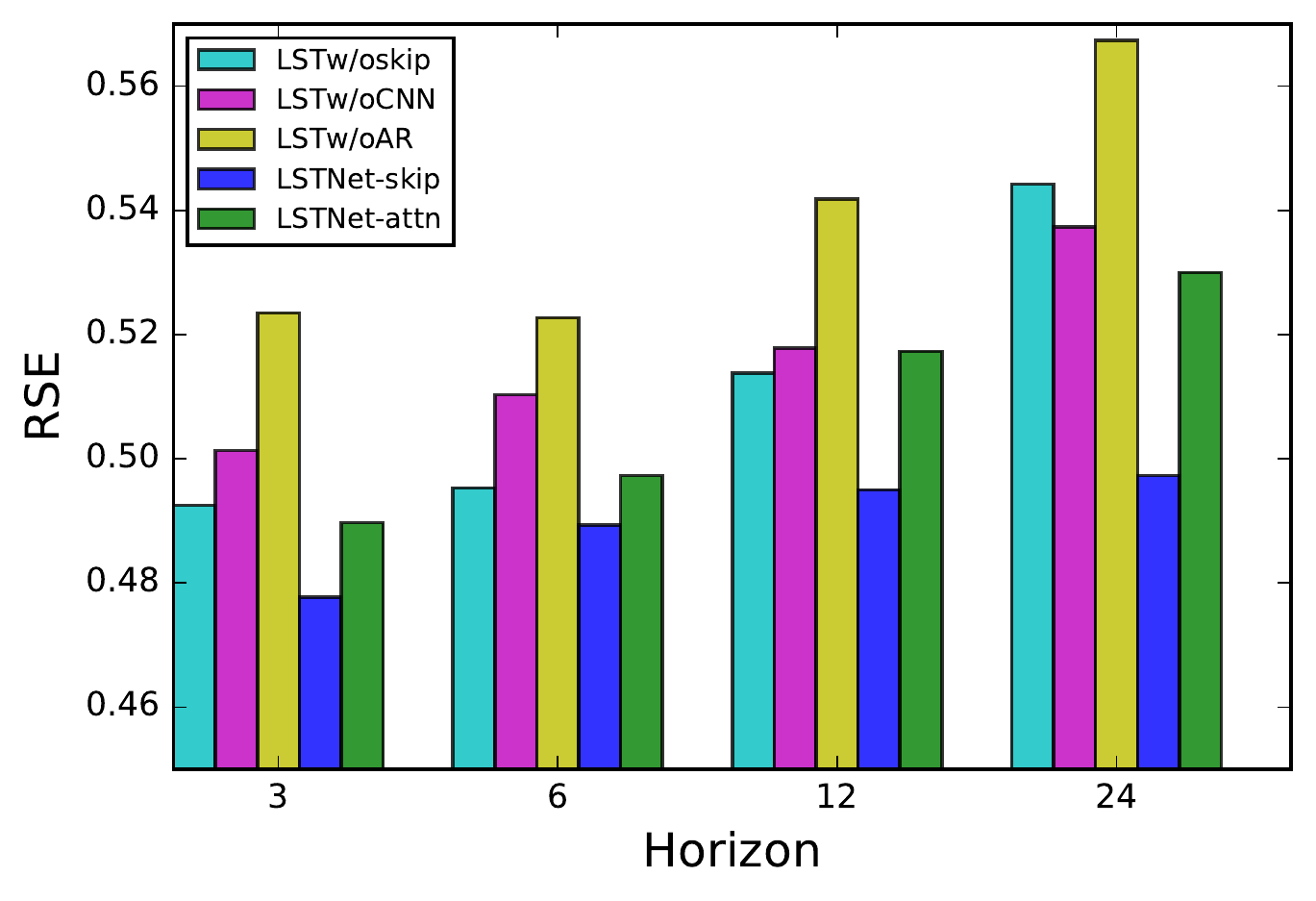}
  \includegraphics[width=.45\linewidth]{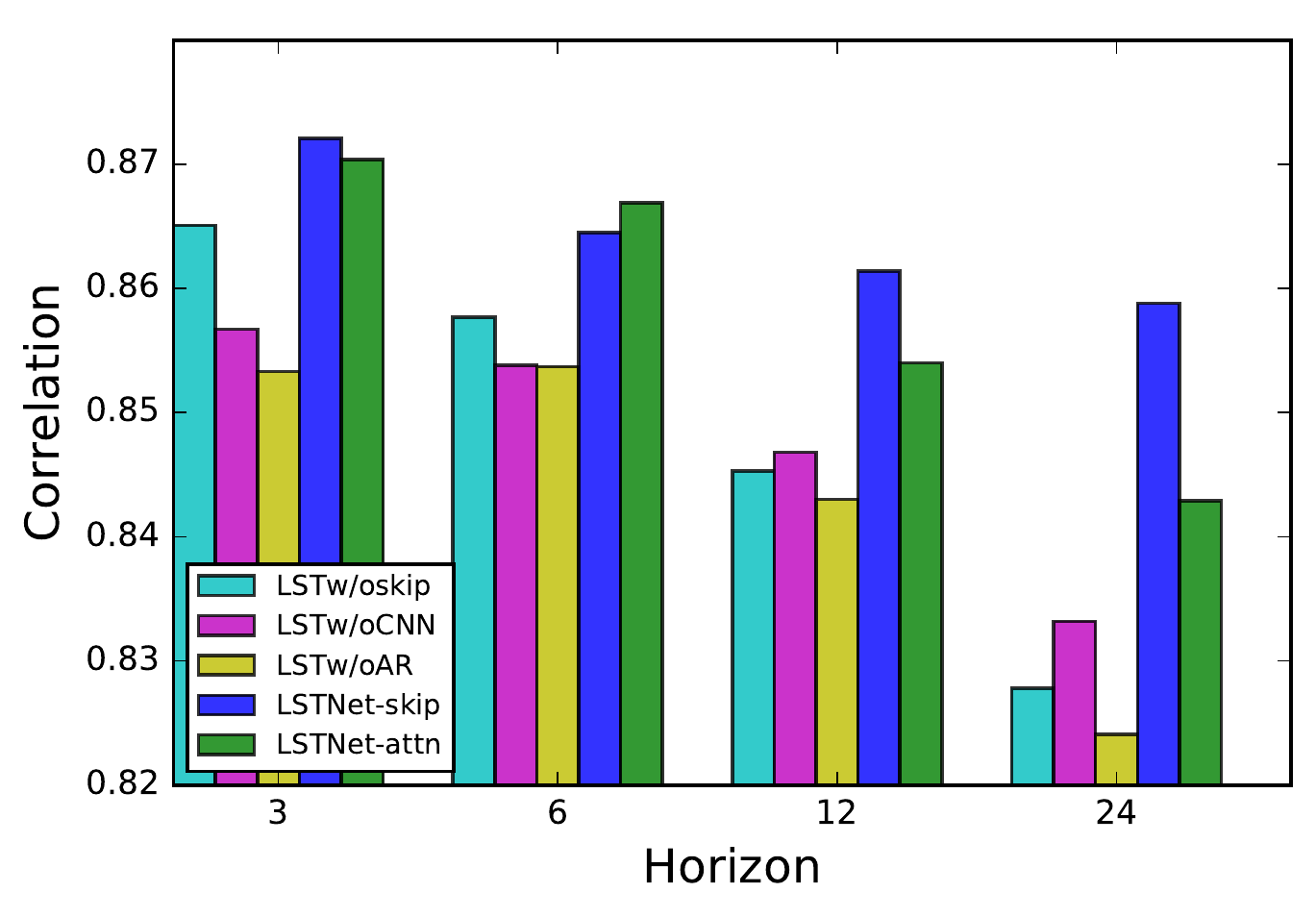}
  \caption{\traffic dataset}
\end{subfigure}

\begin{subfigure}{\textwidth}
  \centering
  \includegraphics[width=.45\linewidth]{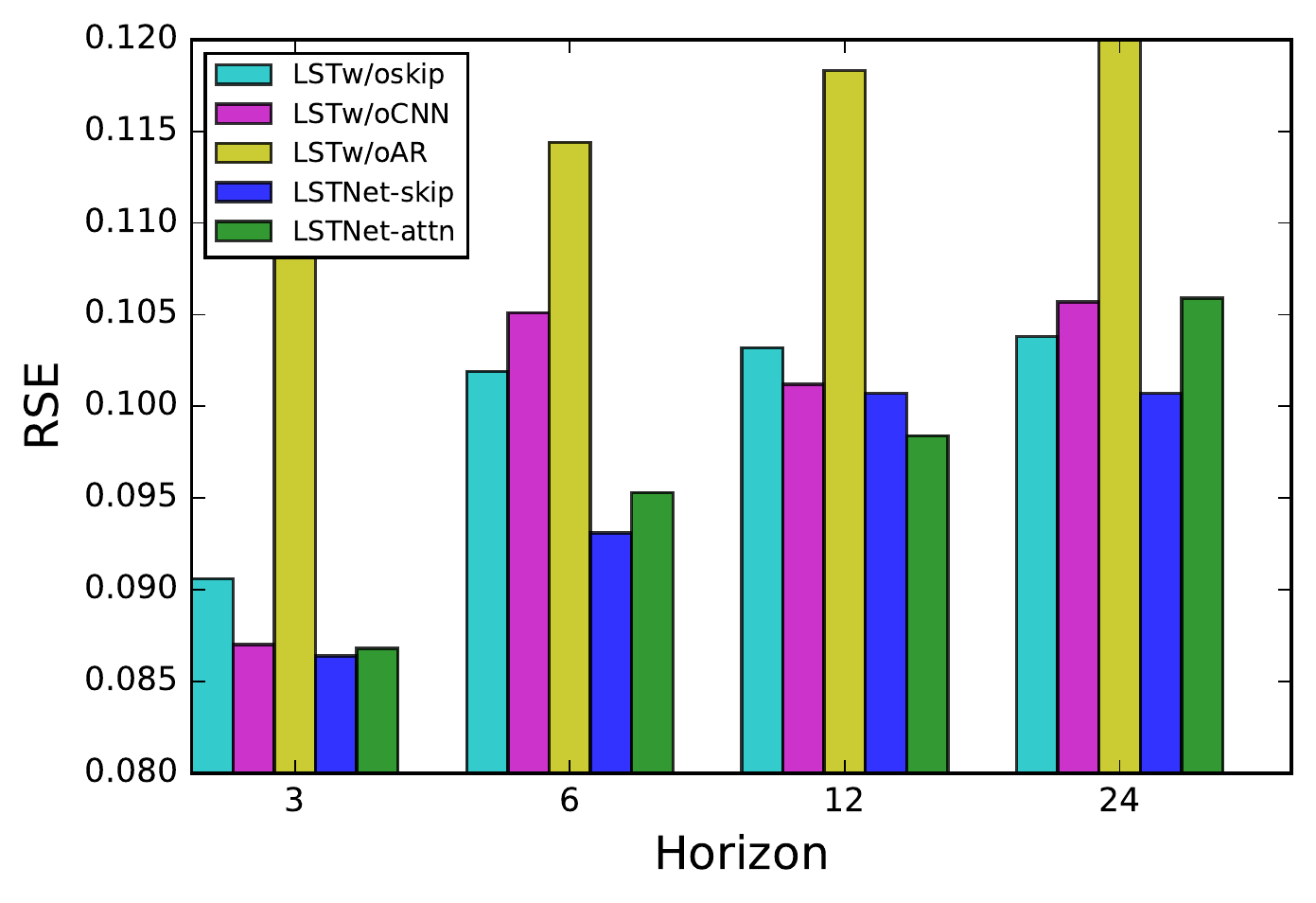}
  \includegraphics[width=.45\linewidth]{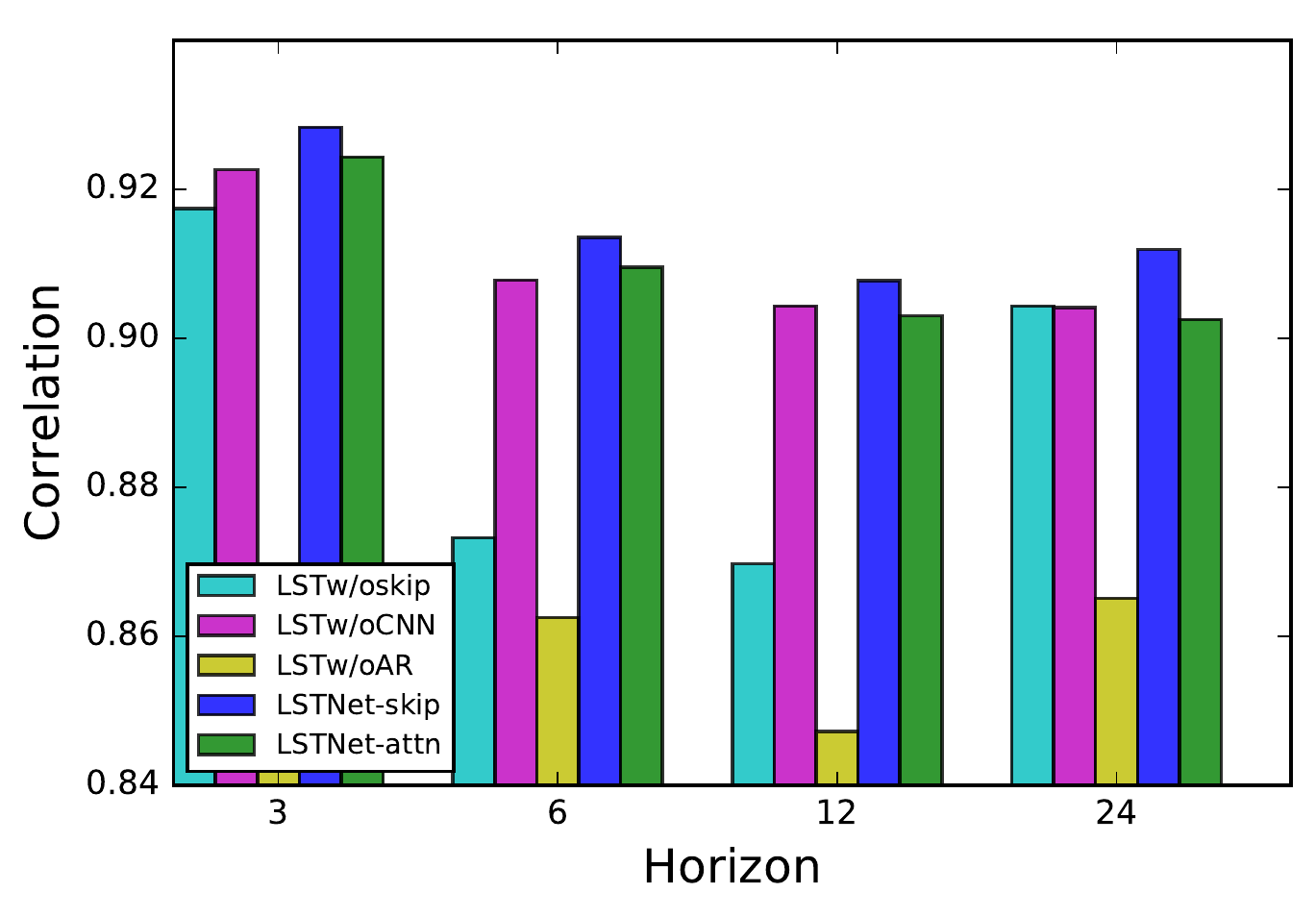}
  \caption{\electricity dataset}
\end{subfigure}
\caption{Results of LSTNet in the ablation tests on the \solar, \traffic and \electricity dataset}
\label{fig:ablation}
\end{figure*}

The test results measured using RSE and CORR are shown in Figure \ref{fig:ablation}\footnote{We omit the results in RAE as it shows similar comparison with respect to the relative performance among the methods.}. Several observations from these results are worth highlighting:
\begin{itemize}
\item The best result on each dataset is obtained with either LST-Skip or LST-Attn.
\item Removing the AR component (in LSTw/oAR) from the full model caused the most significant performance drops on most of the datasets, showing the crucial role of the AR component in general.
\item Removing the Skip and CNN components in (LSTw/oCNN or LSTw/oskip) caused big performance drops on some datasets but not all. All the components of LSTNet together leads to the robust performance of our approach on all the datasets.
\end{itemize}



The conclusion is that our architecture design is most robust across all experiment settings, especially with the large horizons.

\begin{figure*}[!t]
\begin{subfigure}{.45\textwidth}
  \centering
  \includegraphics[width=\linewidth]{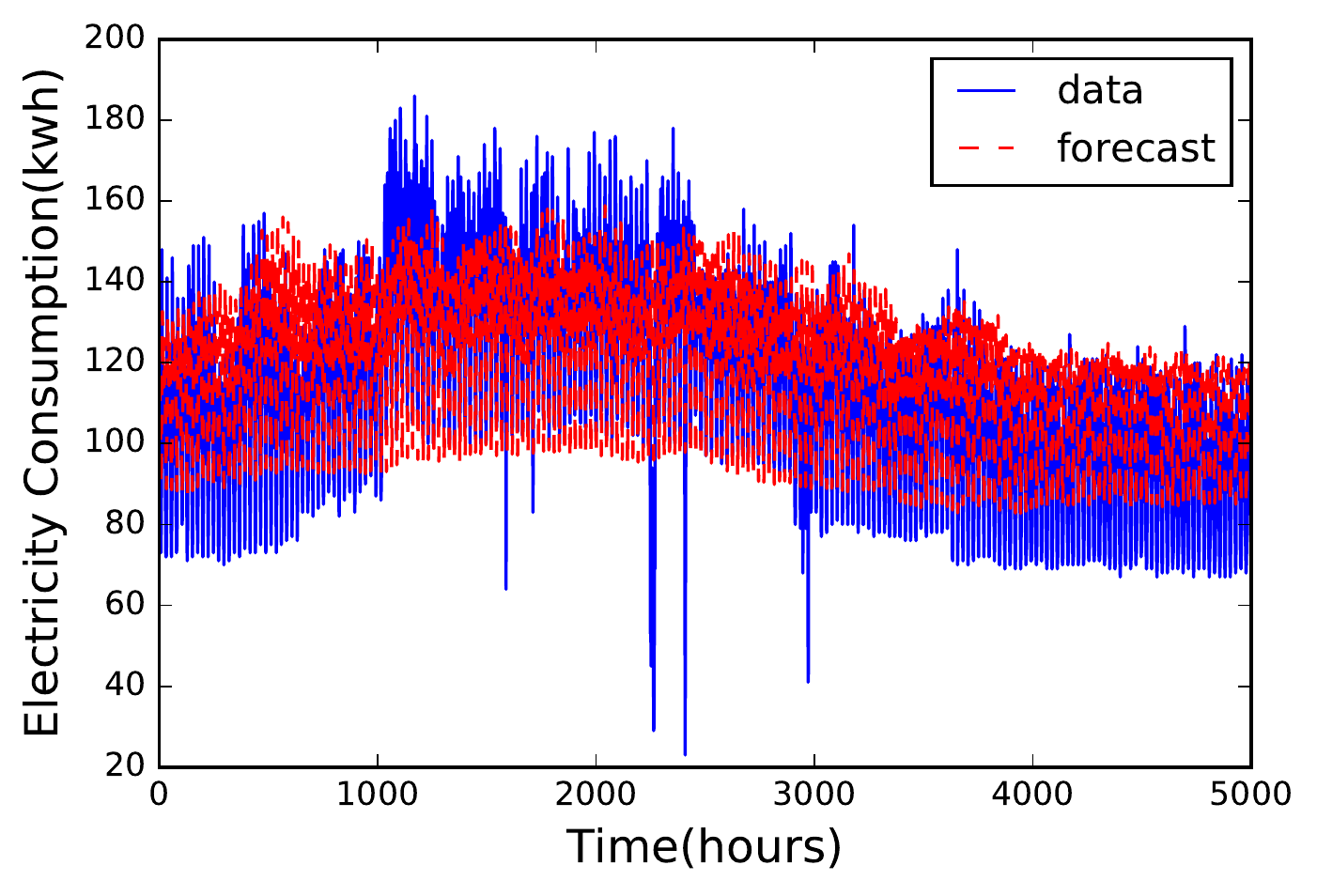}
  \caption{}
\end{subfigure}
\begin{subfigure}{.45\textwidth}
  \centering
  \includegraphics[width=\linewidth]{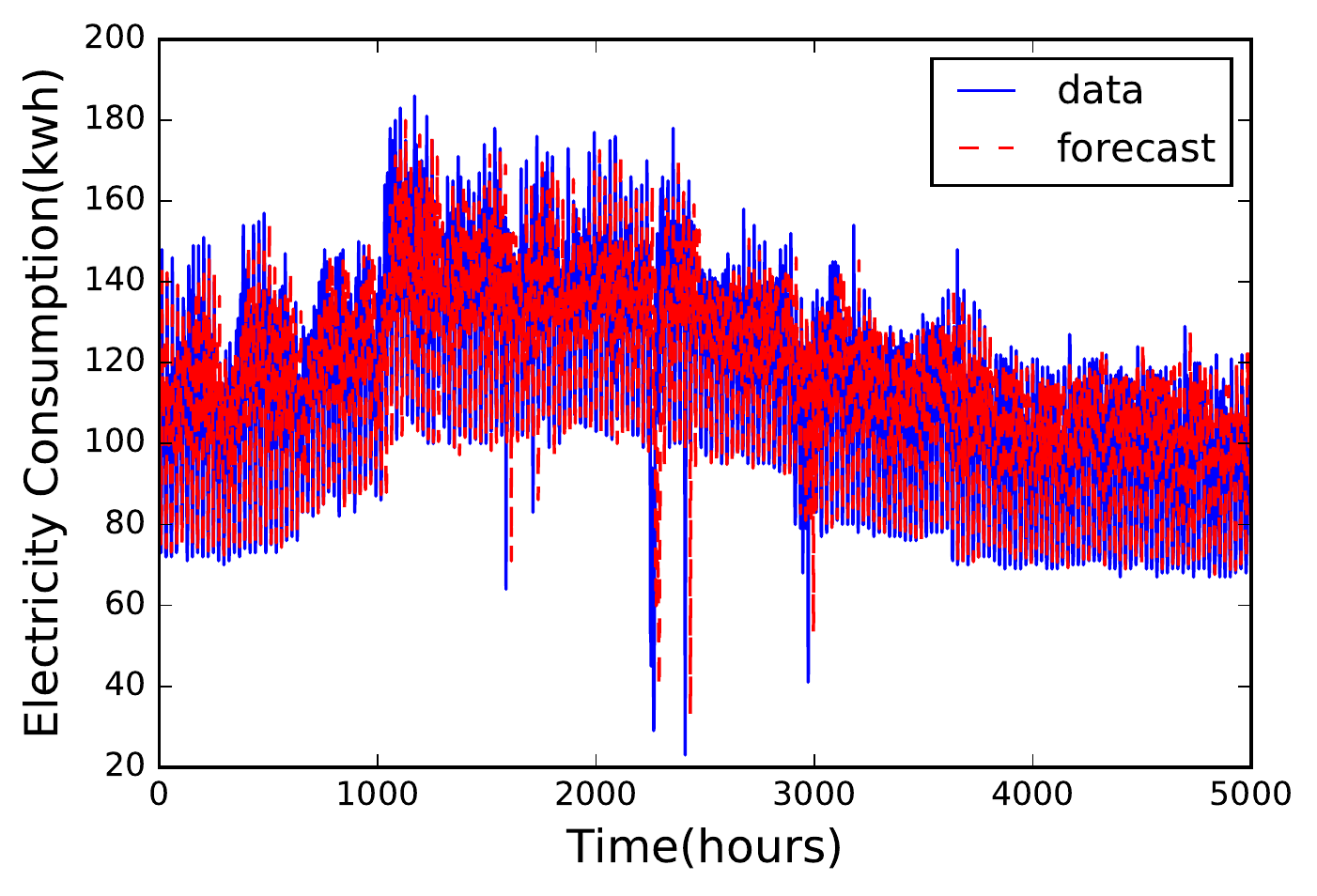}
  \caption{}
\end{subfigure}
\caption{The predicted time series (red) by LSTw/oAR (a) and by LST-Skip  (b) vs. the true data (blue) on \electricity dataset with $horizon = 24$}
\label{fig:electricity}
\end{figure*}

As for why the AR component would have such an important role, our interpretation is that AR is generally robust to the scale changing in data. To empirically validate this intuition we plot one dimension (one variable) of the time series signals in the electricity consumption dataset for the duration from 1 to 5000 hours in Figure \ref{fig:electricity}, where the blue curve is the true data and the red curve is the system-forecasted signals. We can see that the true consumption suddenly increases around the 1000th hour, and that LSTNet-Skip successfully captures this sudden change but LSTw/oAR fails to react properly. 

In order to better verify this assumption, we conduct a simulation experiment. First, we randomly generate an autoregressive process with the scale changing by the following steps. Firstly, we randomly sample a vector, $w \sim N(0, I), w \in \R^{p}$, where $p$ is a given window size. Then the generated autoregressive process $x_t$ can be described as 
\begin{equation}
x_t = \sum_{i=1}^{p} w_ix_{t-i} + \epsilon
\end{equation}
where $\epsilon \sim N(\mu, 1)$. To inject the scale changing, we increase the mean of Gaussian noise by $\mu_0$ every $T$ timestamp. Then the Gaussian noise of time series $x_t$ can be written as 
\begin{equation}
\begin{aligned}
\epsilon \sim N(\left \lfloor{t/T}\right \rfloor \mu_0, 1)
\end{aligned}
\end{equation}
where the $ \left \lfloor{\cdot} \right \rfloor$ denotes the floor function. We split the time series as the training set and test in chronological order, and test the RNN-GRU and the LSTNet models. 
\begin{figure}
  \centering
  \includegraphics[width=\linewidth]{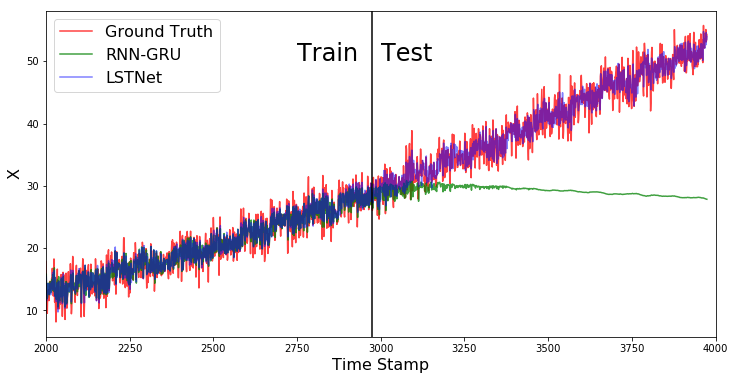}
  \label{fig:tra-var}
\caption{Simulation Test: Left side is the training set and right side is test set.}
\label{fig:sim}
\end{figure}
The result is illustrated in Figure \ref{fig:sim}. Both RNN and LSTNet can memorize the pattern in training set (left side). But, the RNN-GRU model cannot follow the scale changing pattern in the test set (right side). Oppositely, the LSTNet model fits the test set much better. In other words, the normal RNN module, or says the neural-network component in LSTNet, may not be sufficiently sensitive to violated scale fluctuations in data (which is typical in \electricity data possibly due to random events for public holidays or temperature turbulence, etc.), while the simple linear AR model can make a proper adjustment in the forecasting. 

In summary, this ablation study clearly justifies the efficiency of our architecture design. All components have contributed to the excellent and robust performance of LSTNet.

\subsection{Mixture of long- and short-term patterns}
\label{sec:mixture}
\begin{figure*}[!ht]
\begin{subfigure}{.45\textwidth}
  \centering
  \includegraphics[width=\linewidth]{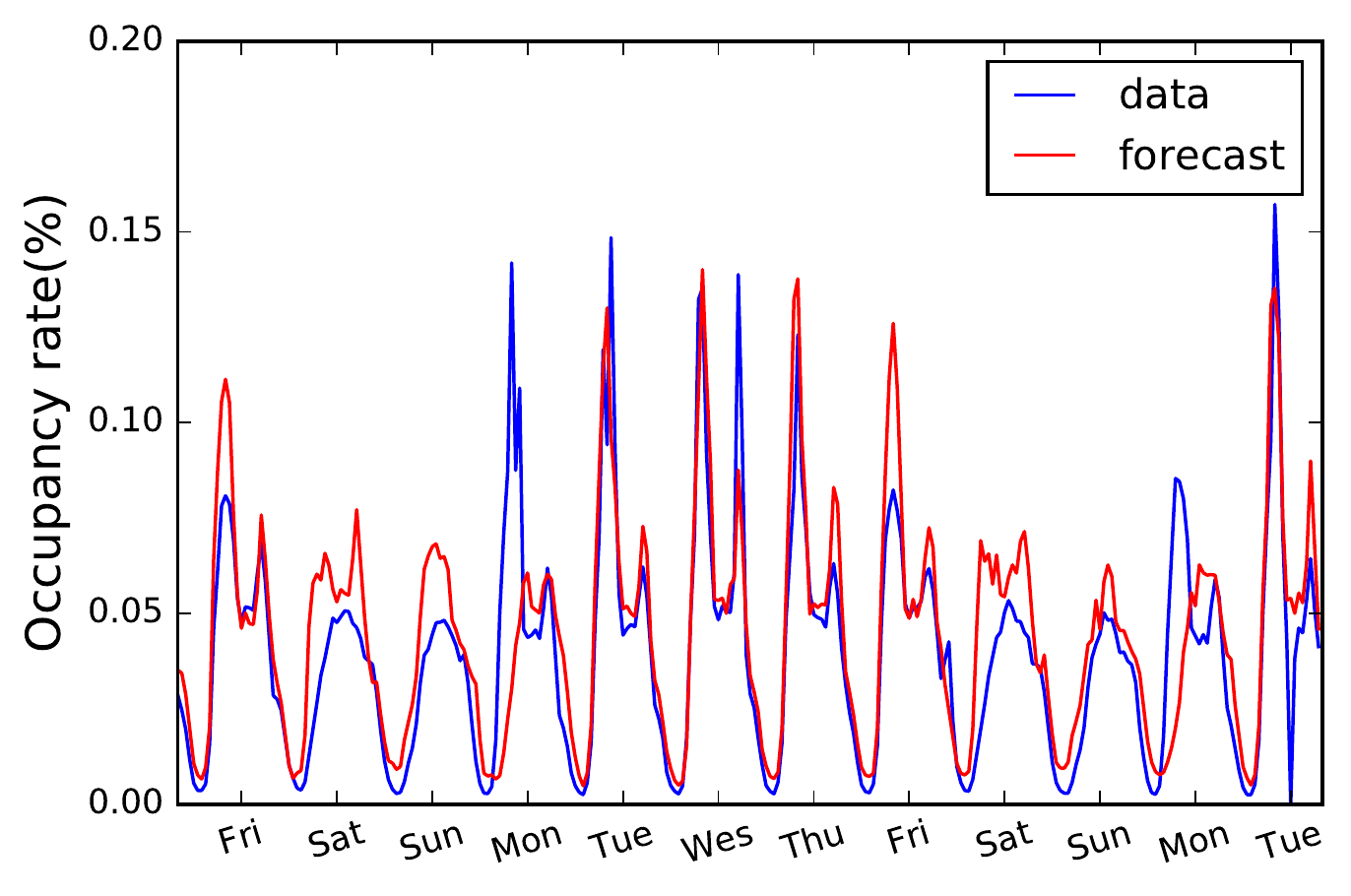}
  \caption{}
  \label{fig:tra-var}
\end{subfigure}
\begin{subfigure}{.45\textwidth}
  \centering
  \includegraphics[width=\linewidth]{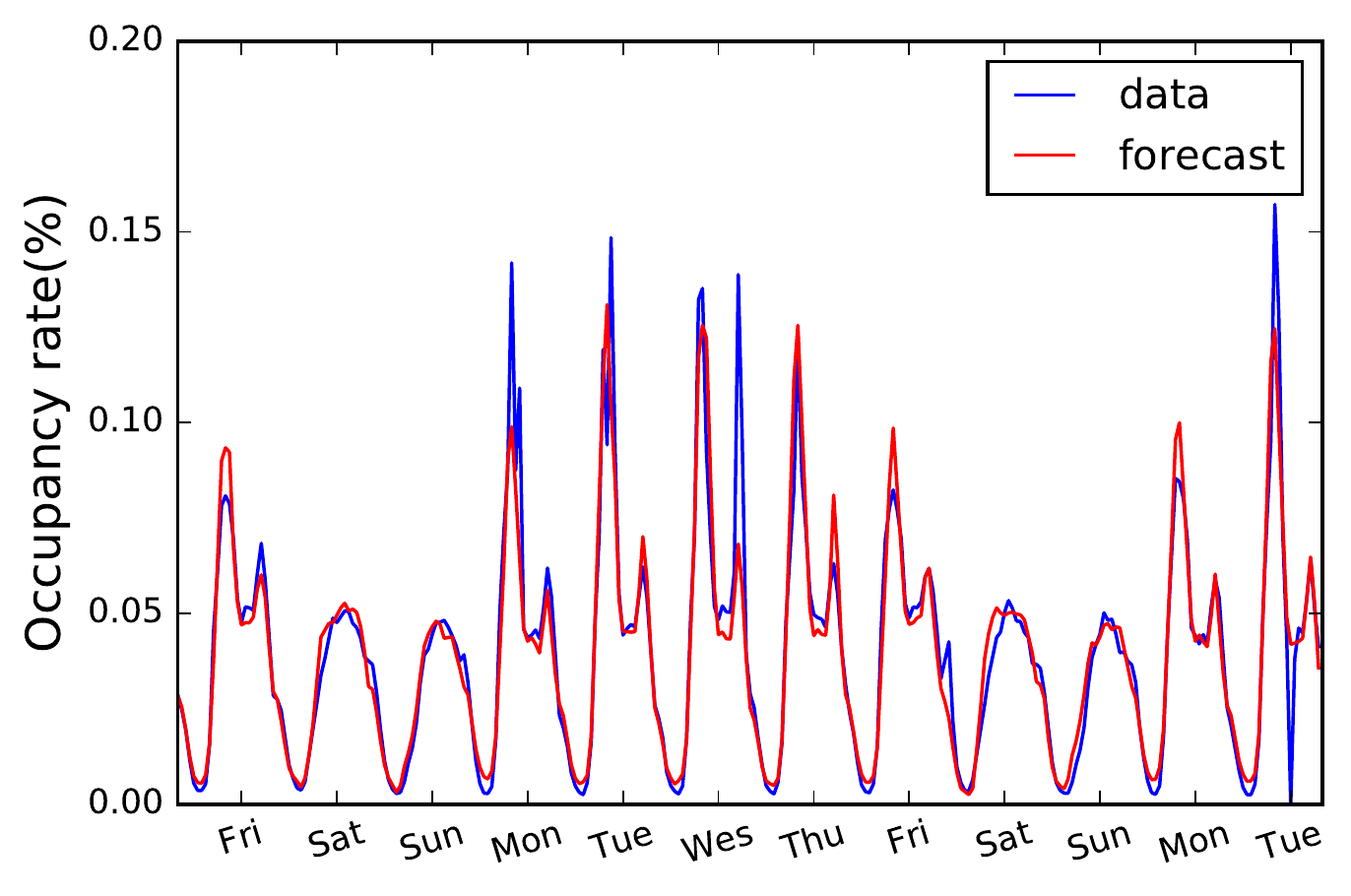}
  \caption{}
  \label{fig:tra-tnn}
\end{subfigure}

\caption{The true time series (blue) and the predicted ones (red) by VAR (a) and by LSTNet (b) for one variable in the \traffic occupation dataset. The X axis indicates the week days and the forecasting $horizon = 24$.  VAR inadequately predicts similar patterns for Fridays and Saturdays, and ones for Sundays and Mondays, while LSTNet successfully captures both the daily and weekly repeating patterns.}
\label{fig:traffic}
\end{figure*}

To illustrate the success of LSTNet in modeling the mixture of short-term and long-term recurring patterns in time series data, Figure \ref{fig:traffic} compares the performance of LSTNet and VAR on an specific time series (one of the output variables) in the \traffic dataset.  As discussed in Section 
\ref{sec:data}, the \traffic data exhibit two kinds of repeating patterns, i.e. the daily ones and the weekly ones. We can see in Figure \ref{fig:traffic} that the true patterns (in blue) of traffic occupancy are very different on Fridays and Saturdays, and another on Sunday and Monday.
The Figure \ref{fig:traffic} is the prediction result of the VAR model (part (a)) and LSTNet (part (b)) of a traffic flow monitor sensor, where their hyper-parameters are chosen according to the RMSE result on the validation set. The figure shows that the VAR model is only capable to deal with the short-term patterns. The pattern of prediction results of the VAR model only depend on the day before the predictions. We can clearly see that the results of it in Saturday (2rd and 9th peaks) and Monday (4th and 11th peaks) is different from the ground truth, where the ground truth of Monday (weekday) has two peaks, one peak for Saturday (weekend). In the contrary, our proposed LSTNet model performs two patterns for weekdays and weekends respectfully. This example proves the ability of LSTNet model to memorize short-term and long-term recurring patterns simultaneously, which the traditional forecasting model is not equipped, and it is crucial in the prediction task of the real world time series signals. 

\section{Conclusion}
\label{sec:conclusion}
In this paper, we presented a novel deep learning framework (LSTNet) for the task of multivariate time series forecasting. By combining the strengths of convolutional and recurrent neural networks and an autoregressive component, the proposed approach significantly improved the state-of-the-art results in time series forecasting on multiple benchmark datasets. With in-depth analysis and empirical evidence, we show the efficiency of the architecture of LSTNet model, and that it indeed successfully captures both short-term and long-term repeating patterns in data, and combines both linear and non-linear models for robust prediction.  

For future research, there are several promising directions in extending the work. Firstly, the skip length $p$ of the skip-recurrent layer is a crucial hyper-parameter. Currently, we manually tune it based on the validation dataset. How to automatically choose $p$ according to data is an interesting problem. Secondly, in the convolution layer we treat each variable dimension equally, but in the real world dataset, we usually have rich attribute information. Integrating them into the LSTNet model is another challenging problem.  
%
\bibliographystyle{abbrv}
\bibliography{sigproc}  
%
%

\end{document}